\documentclass[11pt]{article}

\usepackage[preprint]{acl}

\usepackage{times}
\usepackage{latexsym}

\usepackage[T1]{fontenc}
\usepackage[utf8]{inputenc}

\usepackage{microtype}

\usepackage{inconsolata}

\usepackage{graphicx}

\usepackage{microtype}
\usepackage{inconsolata}
\usepackage{graphicx}
\usepackage{enumitem}
\usepackage{xcolor}
\usepackage{soul}
\usepackage{multirow}
\usepackage{array}
\usepackage{lscape}
\usepackage{svg}
\usepackage{tcolorbox}
\usepackage{amsmath}
\usepackage{subcaption}
\usepackage{lineno}
\usepackage{longtable}
\usepackage{caption}
\usepackage{subcaption}
\usepackage{xurl}
\usepackage{amssymb}
\usepackage{booktabs}  % für toprule, midrule, bottomrule
\usepackage{xcolor}    % für Farben
\usepackage{colortbl}  % für rowcolor

\usepackage{booktabs}        % Für schöne Tabellenlinien (\toprule, \midrule, \bottomrule)
\usepackage{tabularx}        % Für flexiblere Tabellenbreiten (optional)
\usepackage{adjustbox}       % Für Feinjustierungen der Tabellengröße (optional)

\usepackage{siunitx}
\usepackage[table]{xcolor}

\title{nchellwig at SemEval-2026 Task 3: Self-Consistent Structured Generation (SCSG) for Dimensional Aspect-Based Sentiment Analysis using Large Language Models}

\author{
Nils Constantin Hellwig$^{1}$,
Jakob Fehle$^{1}$,
Udo Kruschwitz$^{2}$,
Christian Wolff$^{1}$ \\
\\
$^{1}$Media Informatics Group, University of Regensburg, Regensburg, Germany \\
$^{2}$Information Science Group, University of Regensburg, Regensburg, Germany \\
\\
\texttt{\{nils-constantin.hellwig,jakob.fehle,udo.kruschwitz,christian.wolff\}@ur.de}
}

\begin{document}
\maketitle

\begin{abstract}\sloppy\hbadness=10000
We present \textbf{\underline{S}elf-\underline{C}onsistent \underline{S}tructured \underline{G}eneration (SCSG)} for Dimensional Aspect-Based Sentiment Analysis in SemEval-2026 Task 3 (Track A). SCSG enhances prediction reliability by executing a LoRA-adapted large language model multiple times per instance, retaining only tuples that achieve a majority consensus across runs. To mitigate the computational overhead of multiple forward passes, we leverage vLLM's PagedAttention mechanism for efficient key--value cache reuse. Evaluation across 6 languages and 8 language--domain combinations demonstrates that self-consistency with 15 executions yields statistically significant improvements over single-inference prompting, with our system (leveraging Gemma 3) ranking in the top seven across all settings, achieving second place on three out of four English subsets and first place on Tatar--Restaurant for DimASTE.
\end{abstract}
% Keine externen Daten
% Recheneffektiv durch vllm features
% Subtasks nennen und Semeval task
% Vorteil: Nur domain datensatz, kein Reasoning erforderlich
% max performance with low energy use
% TODOD: Wieso vllm so effektiv
% Vorteil: Open Source

% welche performance im task
% welche performance bei leaderboard
% mistral noch besser
\section{Introduction}

% Was ist ABSA
% Was ist DimABSA / VA
% Was waren bei ABSA bisher für methoden gängig für tuple prediction tasks
% Nennen das Potential von LLMs groß -> smid paper

Aspect-Based Sentiment Analysis (ABSA) extracts and analyzes opinions toward specific aspects within text \citep{zhang2022survey}. ABSA identifies up to four sentiment elements: (1) aspect term, (2) aspect category, (3) opinion term, and (4) sentiment polarity. For instance, from ``The pasta was delicious, but the service was slow'', ABSA extracts (\textit{pasta}, \texttt{FOOD}, \textit{delicious}, positive) and (\textit{service}, \texttt{SERVICE}, \textit{slow}, negative). Aspect Sentiment Quad Prediction (ASQP), which jointly extracts all four elements, represents the most challenging ABSA task \citep{zhang2022survey}.

Research primarily considered generative encoder-decoder architectures \citep{zhang-etal-2021-towards-generative, zhang-etal-2021-aspect-sentiment, hu-etal-2022-improving-aspect, gou-etal-2023-mvp}. Until recently, approaches predominantly leveraged Google's T5-base models \citep{raffel2020exploring} with 220 million parameters as their foundation. These approaches primarily differed in their representation strategies for output tuples, e.g., with sentiment elements being encoded either as structured tuples \citep{zhang-etal-2021-towards-generative} or through paraphrased natural language text \citep{zhang-etal-2021-aspect-sentiment}. The best performance scores when using T5-base were achieved through Multi-view Prompting (MvP) \citep{gou-etal-2023-mvp}, which considers multiple permutations of sentiment element orderings within tuples for both training and inference, thereby enhancing prediction stability by considering diverse positional configurations rather than a single fixed arrangement.

\begin{table}[t]
\centering
\small
\begin{tabular}{p{0.95\columnwidth}}
\toprule
\textbf{Input Text} \\
\midrule
\textit{Decor is nice though service can be spotty.} \\
\midrule
\textbf{DimASQP Output (Quadruplets)} \\
\midrule
$\langle$\textit{Decor}, \textsc{ambience\#general}, \textit{nice}, $v{=}7.00, a{=}7.17$$\rangle$ \\
$\langle$\textit{service}, \textsc{service\#general}, \textit{spotty}, $v{=}7.00, a{=}7.00$$\rangle$ \\
\midrule
\textbf{DimASTE Output (Triplets)} \\
\midrule
$\langle$\textit{Decor}, \textit{nice}, $v{=}7.00, a{=}7.17$$\rangle$ \\
$\langle$\textit{service}, \textit{spotty}, $v{=}7.00, a{=}7.00$$\rangle$ \\
\bottomrule
\end{tabular}
\caption{Example outputs for \textsc{DimASQP} and \textsc{DimASTE} tasks. \textsc{DimASQP} extracts quadruplets containing aspect term, aspect category, opinion term, and valence-arousal ($v, a$) dimensions, where $v, a \in [1,9]$, while \textsc{DimASTE} extracts triplets omitting the aspect category.}
\label{tab:dimabsa_example}
\end{table}

The emergence of substantially more parameter-rich Large Language Models (LLMs) with billions of parameters, pretrained on vast corpora of unlabeled text, enabled further performance gains in ABSA tasks \citep{smid-etal-2024-llama, zhou2024comprehensive}. \citet{smid-etal-2024-llama} reported substantial improvements on the restaurant domain dataset from the SemEval 2016 ABSA shared task \citep{pontiki-etal-2016-semeval}, achieving an F1 score of 78.82 with Orca-2 (13B) compared to 72.76 with MvP on the Target Aspect Sentiment Detection (TASD) task, which extracts aspect terms, aspect categories, and sentiment polarity.

Dimensional ABSA (DimABSA) extends traditional ABSA by replacing categorical polarity labels with continuous valence and arousal dimensions, enabling more nuanced affective characterization \citep{lee2026dimabsa}. In the realm of the SemEval-2026 Task 3 shared task \citep{yu-etal-2026-semeval} (Track A),  two DimABSA subtasks (see Table~\ref{tab:dimabsa_example}) are introduced: Dimensional Aspect Sentiment Triplet Extraction (DimASTE) extracts triplets of aspect term, opinion term, and valence-arousal pairs, while Dimensional Aspect Sentiment Quad Prediction (DimASQP) additionally includes aspect categories.

In this paper, we introduce \textbf{\underline{S}elf-\underline{C}onsistent \underline{S}tructured \underline{G}eneration (SCSG)} for DimABSA. SCSG incorporates self-consistency (SC) prompting as a mechanism to enhance prediction reliability and validity. In SC, each prompt is executed $k$ times with different random seeds, retaining only tuples that achieve a majority consensus across the generated predictions. Since SC requires multiple executions of identical prompts, we leverage vLLM's PagedAttention mechanism for efficient key--value cache reuse, reducing computational overhead. While SC demonstrated substantial performance gains in zero-shot and few-shot prompting settings \citep{hellwig-etal-2025-still}, its application to instruction-tuned LLMs for ABSA has remained unexplored to date. Our key contributions are:

\begin{itemize}
    \item We demonstrate the first application of self-consistency prompting to instruction-tuned LLMs in the ABSA domain, achieving statistically significant improvements across diverse languages and domains.
    \item We develop an efficient inference pipeline using vLLM's PagedAttention for key-value cache reuse across multiple forward passes.
    \item We conduct comprehensive experiments across 6 languages and 8 language--domain combinations. SCSG achieved competitive rankings on the leaderboard, placing in the top seven across all combinations, second place on three of four English subsets, and first place on the Tatar--Restaurant subset for DimASTE.
\end{itemize}

We make our code and results publicly available to facilitate reproducibility and future research.\footnote{\url{https://github.com/NilsHellwig/nchellwig-dimabsa}}

% Guided Decoding als Technik nennen
% Unserer Ansatz: Verschiedene LLMs evaluieren, verschiedene anzahl an epochen -> auf kleinerem Modell
% Nennen welchen Subtask wir machen
% Open Source: GitHub

% TODOD: Nennen, dass er unabhänig funktioniert, keine weiteren Datenquellen oder viele Daten
% TODO: https://semeval.github.io/paper-requirements.html
% TODO: Sagen, dass gemma auch in eine customer gpu passt 
\section{System Overview}

This section provides an in-depth description of our system. 

\subsection{Fine-Tuning Setup \& Prompting Strategy}

SCSG employs parameter-efficient fine-tuning using Low-Rank (LoRA) Adaptation \citep{hu2022lora}. For training, we utilize the Unsloth library\footnote{\url{https://github.com/unslothai/unsloth}}, which enables memory-efficient fine-tuning of LLMs.

\paragraph{Training Hyperparameters}
We applied the hyperparameters employed by \citet{smid-etal-2024-llama}, who fine-tuned LLMs for ABSA tuple prediction tasks. We configured LoRA with rank $r=64$ and scaling factor $\alpha=16$, applying adaptation to both attention and MLP modules while keeping vision layers frozen. We set the dropout rate to 0 and used no bias terms in the adaptation layers. We trained for 5 epochs with a batch size of 16 and set the learning rate to $2 \times 10^{-4}$. The maximum sequence length was set to 1,024 tokens. We applied response-only training, where the loss is computed exclusively on the model response tokens.

\paragraph{Prompt}

As presented in Appendix~\ref{appendix:prompt}, the employed prompt consists of detailed descriptions of the sentiment elements, including valence and arousal, and the desired output format.
The descriptions of aspect term, opinion term, and aspect category follow those used by \citet{hellwig-etal-2025-still}, \citet{gou-etal-2023-mvp}, and \citet{smid-etal-2024-llama} for LLM prompting. For training data in Japanese, Russian, Tatar, Ukrainian, and Chinese, a translated version of the prompt was employed.

\subsection{Validation Module}

SCSG employs a validation pipeline to improve the quality of output tuples. First, we utilize a self-consistency mechanism \citep{wang2022self} that has previously demonstrated improvements in ABSA tasks \citep{hellwig-etal-2025-still, bao-etal-2025-exploring} but has not yet been adapted to instruction-tuned LLMs. Specifically, we execute the LLM $k$ times with temperature $t=0.8$ to better capture model uncertainty and filter tuples for which the model exhibits low confidence. Following the voting mechanisms proposed by \citet{gou-etal-2023-mvp} and \citet{hellwig-etal-2025-still}, a tuple is included in the final prediction if and only if it appears in at least $\lceil k/2 \rceil + 1$ of the $k$ predictions, thereby requiring strict majority agreement.

Two tuples are considered identical if all sentiment elements match, ignoring the valence-arousal pair. For tuples deemed identical under this criterion, the final valence and arousal values are computed as the arithmetic mean across all matching instances. An illustrative example of the calculation is provided in Appendix~\ref{appendix:validation_mechanism}. Finally, we removed tuples containing text spans not present in the input review, capped valence and arousal values to the valid range $[1.00, 9.00]$, and discarded tuples containing aspect categories that were not considered.

\subsection{Inference Optimization}

The computational requirements of generating $k$ predictions per instance necessitate efficient inference optimization. Since identical prompts are executed multiple times, differing only in the specific example to be labeled across language, domain, and subtask combinations, SCSG leverages vLLM \citep{kwon2023vllm}, a high-throughput inference engine for LLMs. Specifically, it leverages vLLM's PagedAttention mechanism to enable efficient key--value (KV) cache reuse across multiple forward passes. In addition, batched inference allows all predictions ($k \times N_{\text{test}}$) to be processed jointly rather than sequentially, where $N_{\text{test}}$ denotes the number of examples in the test set.

%Furthermore, building upon approaches previously employed with T5-based methods, we implement guided decoding using context-free grammars, validating each generated token during decoding. In our context, this ensures that predicted aspect and opinion terms must actually occur in the text, aspect categories must belong to the fixed set of considered categories, and valence and arousal scores must lie between $1.00$ and $9.00$ as defined for the SemEval track, with exactly two decimal places. Since vLLM does not natively support batched processing in combination with example-specific grammars, we implemented a modified version of vLLM\footnote{\url{https://github.com/NilsHellwig/vllm/tree/issue-19007-guided-decoding-params}}. Finally, we removed duplicate tuples in predicted labels as instructed by the task organizers.

% Wieso nicht MvP
%% Prompt
% LLM Begründen mit Kompatibilität von vLLM und unsloth

%% Guded Decoding

\section{Experimental Setup}

\paragraph{Datasets}
We evaluated SCSG on the multilingual DimABSA benchmark introduced in SemEval-2026 Task 3 \citep{lee2026dimabsa}, spanning 6 languages (English, Japanese, Russian, Tatar, Ukrainian, Chinese) and 3 domains (restaurant, laptop, hotel) for both DimASTE and DimASQP. For each task-language-domain configuration, models were fine-tuned on the merged training and validation sets and evaluated on the official test sets. Dataset statistics are provided in Appendix~\ref{appendix:dataset_stats}.

\begin{table*}[ht]
    \centering
    \renewcommand{\arraystretch}{0.9}
    \begin{subtable}{1.0\textwidth}
        \centering
        \resizebox{\textwidth}{!}{%
            \begin{tabular}{@{}ll|lll|lll|lll|lll}
                \toprule
                & & \multicolumn{3}{c|}{\textbf{Baseline}} & \multicolumn{3}{c|}{\textbf{5 Views}} & \multicolumn{3}{c|}{\textbf{10 Views}} & \multicolumn{3}{c}{\textbf{15 Views}} \\
                \cmidrule(lr){3-5} \cmidrule(lr){6-8} \cmidrule(lr){9-11} \cmidrule(lr){12-14}
                \textbf{Language} & \textbf{Domain} & cPrec & cRec & cF1 & cPrec & cRec & cF1 & cPrec & cRec & cF1 & cPrec & cRec & cF1 \\
                \midrule
English & Restaurant & 72.28 & \textbf{67.30} & 69.70 & 73.21 & 66.78 & 69.85 & \textbf{74.30} & 66.40 & \textbf{70.13}***† & 73.36 & 66.78 & 69.92 \\
\rowcolor[gray]{0.96} English & Laptop & 65.58 & \textbf{56.11} & 60.48 & 66.21 & 55.93 & 60.64 & \textbf{67.29} & 55.78 & \textbf{60.99}***†† & 66.52 & 56.10 & 60.87* \\
Japanese & Hotel & 53.41 & 54.53 & 53.96 & 54.84 & 54.09 & 54.46** & \textbf{57.27} & 53.21 & 55.16***†† & 56.07 & \textbf{54.64} & \textbf{55.35}***†† \\
\rowcolor[gray]{0.96} Russian & Restaurant & 52.46 & \textbf{58.32} & 55.24 & 55.11 & 57.18 & 56.13 & \textbf{55.88} & 55.87 & 55.88 & 55.42 & 57.42 & \textbf{56.40} \\
Tatar & Restaurant & 47.47 & \textbf{50.50} & 48.94 & 50.13 & 49.78 & 49.96 & \textbf{52.15} & 48.81 & 50.42 & 51.54 & 50.48 & \textbf{51.00} \\
\rowcolor[gray]{0.96} Ukrainian & Restaurant & 50.28 & \textbf{53.46} & 51.82 & 53.24 & 52.55 & \textbf{52.89} & \textbf{54.44} & 51.20 & 52.77 & 53.32 & 52.08 & 52.69 \\
Chinese & Restaurant & 54.22 & \textbf{55.53} & 54.87 & 55.00 & 54.03 & 54.51 & \textbf{56.37} & 53.43 & 54.86 & 55.60 & 54.22 & \textbf{54.90} \\
\rowcolor[gray]{0.96} Chinese & Laptop & 48.32 & \textbf{49.96} & 49.13 & 51.55 & 48.93 & 50.20*** & \textbf{53.45} & 48.05 & 50.61***† & 52.58 & 49.24 & \textbf{50.86}***†† \\
\midrule
\textbf{Average} &  & 55.50 & \textbf{55.71} & 55.52 & 57.41 & 54.91 & 56.08 & \textbf{58.89} & 54.09 & 56.35 & 58.05 & 55.12 & \textbf{56.50} \\
\bottomrule
            \end{tabular}%
        }
        \caption{Subtask 2: \textsc{DimASTE}}
    \end{subtable}

    \vspace{0.3em}

    \begin{subtable}{1.0\textwidth}
        \centering
        \renewcommand{\arraystretch}{0.9}
        \resizebox{\textwidth}{!}{%
            \begin{tabular}{@{}ll|lll|lll|lll|lll}
                \toprule
                & & \multicolumn{3}{c|}{\textbf{Baseline}} & \multicolumn{3}{c|}{\textbf{5 Views}} & \multicolumn{3}{c|}{\textbf{10 Views}} & \multicolumn{3}{c}{\textbf{15 Views}} \\
                \cmidrule(lr){3-5} \cmidrule(lr){6-8} \cmidrule(lr){9-11} \cmidrule(lr){12-14}
                \textbf{Language} & \textbf{Domain} & cPrec & cRec & cF1 & cPrec & cRec & cF1 & cPrec & cRec & cF1 & cPrec & cRec & cF1 \\
\midrule
English & Restaurant & 66.64 & \textbf{60.23} & 63.27 & 67.93 & 59.76 & 63.59** & \textbf{68.93} & 59.52 & 63.88***†† & 68.49 & 60.00 & \textbf{63.97}***††† \\
\rowcolor[gray]{0.96} English & Laptop & 42.60 & \textbf{36.70} & 39.43 & 44.62 & 35.25 & 39.39 & \textbf{47.32} & 34.81 & 40.11**†† & 46.23 & 35.67 & \textbf{40.27}***†† \\
Japanese & Hotel & 37.54 & \textbf{38.94} & 38.23 & 40.89 & 37.64 & 39.20** & \textbf{43.70} & 36.97 & 40.06***† & 43.00 & 38.03 & \textbf{40.37}***†† \\
\rowcolor[gray]{0.96} Russian & Restaurant & 48.42 & \textbf{51.03} & 49.69 & 50.79 & 50.49 & 50.64** & \textbf{52.35} & 49.40 & 50.83*** & 51.31 & 50.63 & \textbf{50.96}*** \\
Tatar & Restaurant & 43.33 & \textbf{45.72} & 44.49 & 47.56 & 44.76 & \textbf{46.12}*** & \textbf{49.00} & 43.08 & 45.85*** & 47.77 & 44.16 & 45.90*** \\
\rowcolor[gray]{0.96} Ukrainian & Restaurant & 45.67 & \textbf{46.13} & 45.90 & 48.21 & 45.55 & 46.84* & \textbf{50.31} & 44.61 & 47.29*** & 49.67 & 45.68 & \textbf{47.59}***† \\
Chinese & Restaurant & 47.83 & \textbf{49.52} & 48.66 & 49.99 & 49.06 & 49.52*** & \textbf{51.42} & 48.19 & \textbf{49.75}***† & 50.49 & 48.90 & 49.68*** \\
\rowcolor[gray]{0.96} Chinese & Laptop & 38.30 & \textbf{39.94} & 39.10 & 41.40 & 38.16 & 39.71** & \textbf{44.04} & 36.76 & 40.07*** & 42.59 & 38.13 & \textbf{40.24}***† \\
\midrule
\textbf{Average} &  & 46.29 & \textbf{46.03} & 46.10 & 48.92 & 45.09 & 46.88 & \textbf{50.88} & 44.17 & 47.23 & 49.94 & 45.15 & \textbf{47.37} \\
\bottomrule
            \end{tabular}%
        }
        \caption{Subtask 3: \textsc{DimASQP}}
    \end{subtable}
    \caption{Performance on the test set for \textsc{DimASTE} and \textsc{DimASQP}: Comparison of vanilla prompting (Baseline) and self-consistency (SC) with 5, 10, or 15 prompt executions. Results show continuous-level precision (cPrec), recall (cRec), and F1-score (cF1) in \%. \textbf{Bold} values indicate the best performance for each language–domain pair and metric. Asterisks denote statistical significance of the improvement over the baseline, while daggers ($\dagger$) and double daggers ($\ddagger$) denote significance over 5 and 10 views, respectively, based on Holm-Bonferroni corrected p-values ($*: p < 0.05, **: p < 0.01, ***: p < 0.001$).}
    \label{tab:test_performance_combined}
\end{table*}

\paragraph{Training Configuration} 
All experiments were conducted on an NVIDIA RTX Pro 6000 (Blackwell generation) GPU with 96~GB of VRAM. We evaluated the 27B parameter variant of Gemma 3 \citep{gemmateam2025gemma3technicalreport} using 4-bit quantization\footnote{\url{https://huggingface.co/unsloth/gemma-3-27b-it-unsloth-bnb-4bit}}, which provides compatibility with both vLLM and Unsloth frameworks. Following the submission phase, we additionally evaluated SCSG on Mistral-Small-3.2 (24B)\footnote{\url{https://huggingface.co/unsloth/Mistral-Small-3.2-24B-Instruct-2506-bnb-4bit}} and Qwen3 (32B) \citep{Yang2025-ec}\footnote{\url{https://huggingface.co/unsloth/Qwen3-32B-unsloth-bnb-4bit}} to assess generalization across models developed by different institutions in different regions (France and China, respectively).

\paragraph{Evaluation}

Following the evaluation protocol defined by the task organizers, we used continuous F1 (cF1) as the primary metric. cF1 is based on the continuous true positive (cTP). A prediction is considered (partially) correct only if all categorical elements exactly match the gold annotation, i.e., $(A, O)$ for DimASTE triplets or $(A, C, O)$ for DimASQP quadruplets. Each such prediction is assigned an initial true positive score of 1, which is reduced according to the normalized valence--arousal (VA) prediction error:

\begin{equation}
    \mathrm{cTP}(t) = 1 - \frac{(V_p - V_g)^2 + (A_p - A_g)^2}{D_{\max}},
\end{equation}

where $(V_p, A_p)$ and $(V_g, A_g)$ denote the predicted and gold VA values, respectively, and $D_{\max} = 8^2 + 8^2 = 128$ is the maximum possible squared Euclidean distance in the VA space on the $[1,9]$ scale. Predictions with no categorical match receive a cTP score of 0.

Based on aggregated cTP values, continuous Precision (cPrec) and Recall (cRec) are computed analogously to their standard counterparts, with the numerator given by the sum of cTP values across all predictions. The continuous F1 score (cF1) is defined as the harmonic mean of cPrec and cRec. When VA predictions are perfect (i.e., $\mathrm{dist}=0$), cF1 reduces to the standard F1 score. 

We executed the LLM five times with different random seeds (0-4) and computed the average across all runs to provide robust results. For a given configuration, the total number of predictions amounts to $n_{\text{seeds}} \times k \times N_{\text{test}}$. We submitted the predictions corresponding to the first seed and release all predicted labels to facilitate future metric computation and reproducibility.

\paragraph{Pilot study: Determining the Optimal Output Validation Mechanism} 
Prior to our submission for the task leaderboard, we assessed four values of $k$ for the self-consistency (SC) mechanism: $k \in \{1, 5, 10, 15\}$, where $k=1$ corresponds to vanilla prompting with a single execution. We used the training set to fine-tune the models and evaluated it on the development set.

%% Allgemein
% Welche Ansätze: Einzeln vs. multilanguage

%% Hyperparametertuning
% Welche Splits 80 vs 20

%% Welche Metriken
% welche gpu
% cross validation nennen

% 1. vs. 2 bei mehreren sprachen
% 5 vs 10 vs 15 bei mehreren

% Was möchte ich vergleichen nennen? 
% nicht nur ein ansatz submitten - system soll unabhängig funktionieren ohne wissen zu anderen subsets

% TODO: irgendwo nennen, dass wen SC false, dann temp=0.0
\section{Results}

\begin{table*}[t]
\centering
\small

\begin{subtable}{\textwidth}
\centering
\resizebox{\textwidth}{!}{%
\begin{tabular}{@{}llrrrrrrr@{}}
\toprule
\textbf{Language} & \textbf{Domain} & \textbf{1st} & \textbf{2nd} & \textbf{3rd} & \textbf{4th} & \textbf{5th} & \textbf{6th} & \textbf{7th} \\
\midrule
English & Laptop & \cellcolor[HTML]{dfefb2} Takoyaki (63.66) & \cellcolor[HTML]{e7f3b2} PALI (62.42) & \cellcolor[HTML]{edf6b2} PAI (61.69) & \cellcolor[HTML]{f2f8b2} \textbf{nchellwig (60.92)} & \cellcolor[HTML]{f6ecb2} SokraTUM (56.35) & \cellcolor[HTML]{f5ebb2} ICT-NLP (56.22) & \cellcolor[HTML]{f3e5b2} EmberAI (55.46) \\
English & Restaurant & \cellcolor[HTML]{b2d8b2} Takoyaki (70.21) & \cellcolor[HTML]{b4dab2} \textbf{nchellwig (69.85)} & \cellcolor[HTML]{b8dbb2} PALI (69.28) & \cellcolor[HTML]{badcb2} PAI (69.03) & \cellcolor[HTML]{c8e3b2} kevinyu66 (67.07) & \cellcolor[HTML]{cce5b2} EmberAI (66.41) & \cellcolor[HTML]{d5eab2} AILS-NTUA (65.18) \\
Japanese & Hotel & \cellcolor[HTML]{fcf9b2} TeleAI (58.37) & \cellcolor[HTML]{f8f0b2} TeamLasse (56.94) & \cellcolor[HTML]{f7efb2} PAI (56.82) & \cellcolor[HTML]{f7eeb2} PALI (56.66) & \cellcolor[HTML]{f2e4b2} \textbf{nchellwig (55.18)} & \cellcolor[HTML]{edd9b2} kevinyu66 (53.66) & \cellcolor[HTML]{edd7b2} Takoyaki (53.40) \\
Russian & Restaurant & \cellcolor[HTML]{fbf6b2} PAI (57.93) & \cellcolor[HTML]{f9f3b2} TeleAI (57.36) & \cellcolor[HTML]{f9f2b2} PALI (57.24) & \cellcolor[HTML]{f6ecb2} \textbf{nchellwig (56.40)} & \cellcolor[HTML]{f3e7b2} Takoyaki (55.64) & \cellcolor[HTML]{f1e2b2} Habib university (54.92) & \cellcolor[HTML]{efdcb2} ALPS-Lab (54.14) \\
Tatar & Restaurant & \cellcolor[HTML]{e6c8b2} \textbf{nchellwig (51.19)} & \cellcolor[HTML]{e5c6b2} Takoyaki (50.92) & \cellcolor[HTML]{dfbab2} PAI (49.08) & \cellcolor[HTML]{deb6b2} TeleAI (48.63) & \cellcolor[HTML]{ddb5b2} Habib university (48.39) & \cellcolor[HTML]{ddb4b2} PALI (48.28) & \cellcolor[HTML]{dcb2b2} ALPS-Lab (47.98) \\
Ukrainian & Restaurant & \cellcolor[HTML]{fbf6b2} PAI (57.87) & \cellcolor[HTML]{f8f1b2} TeleAI (57.12) & \cellcolor[HTML]{f7eeb2} PALI (56.71) & \cellcolor[HTML]{f5eab2} ALPS-Lab (56.13) & \cellcolor[HTML]{f0deb2} Takoyaki (54.38) & \cellcolor[HTML]{ecd6b2} Habib university (53.24) & \cellcolor[HTML]{ebd3b2} \textbf{nchellwig (52.85)} \\
Chinese & Laptop & \cellcolor[HTML]{ecd5b2} PALI (53.08) & \cellcolor[HTML]{ecd5b2} PAI (53.06) & \cellcolor[HTML]{ebd4b2} TeleAI (52.92) & \cellcolor[HTML]{e5c7b2} \textbf{nchellwig (51.10)} & \cellcolor[HTML]{e0bbb2} ALPS-Lab (49.35) & \cellcolor[HTML]{dcb3b2} TeamLasse (48.07) & \cellcolor[HTML]{dcb2b2} kevinyu66 (48.02) \\
Chinese & Restaurant & \cellcolor[HTML]{f6ecb2} PAI (56.38) & \cellcolor[HTML]{f6ebb2} PALI (56.34) & \cellcolor[HTML]{f1e1b2} \textbf{nchellwig (54.88)} & \cellcolor[HTML]{f0dfb2} TeleAI (54.48) & \cellcolor[HTML]{eedab2} Takoyaki (53.82) & \cellcolor[HTML]{ecd6b2} TeamLasse (53.20) & \cellcolor[HTML]{ead1b2} ALPS-Lab (52.47) \\
\bottomrule
\end{tabular}
}
\caption{Subtask 2: \textsc{DimASTE}}
\label{tab:leaderboard-dimaste}
\end{subtable}

\vspace{0.5em}

\begin{subtable}{\textwidth}
\centering
\resizebox{\textwidth}{!}{%
\begin{tabular}{@{}llrrrrrrr@{}}
\toprule
\textbf{Language} & \textbf{Domain} & \textbf{1st} & \textbf{2nd} & \textbf{3rd} & \textbf{4th} & \textbf{5th} & \textbf{6th} & \textbf{7th} \\
\midrule
English & Laptop & \cellcolor[HTML]{f3e5b2} Takoyaki (42.27) & \cellcolor[HTML]{eedbb2} \textbf{nchellwig (40.06)} & \cellcolor[HTML]{ead2b2} PALI (37.93) & \cellcolor[HTML]{ead0b2} PAI (37.58) & \cellcolor[HTML]{e2c0b2} ALPS-Lab (33.95) & \cellcolor[HTML]{e0bbb2} TeleAI (32.81) & \cellcolor[HTML]{dcb2b2} The Classics (30.72) \\
English & Restaurant & \cellcolor[HTML]{b2d8b2} Takoyaki (65.14) & \cellcolor[HTML]{b7dbb2} \textbf{nchellwig (64.03)} & \cellcolor[HTML]{b7dbb2} PALI (63.95) & \cellcolor[HTML]{c0dfb2} ALPS-Lab (62.02) & \cellcolor[HTML]{c9e4b2} AILS-NTUA (59.88) & \cellcolor[HTML]{cce5b2} TeamLasse (59.37) & \cellcolor[HTML]{cfe7b2} HUS@NLP-VNU (58.71) \\
Japanese & Hotel & \cellcolor[HTML]{f3e6b2} PALI (42.52) & \cellcolor[HTML]{f0dfb2} Takoyaki (40.86) & \cellcolor[HTML]{efdcb2} NLANGPROC (40.28) & \cellcolor[HTML]{eedbb2} TeamLasse (39.92) & \cellcolor[HTML]{eedab2} \textbf{nchellwig (39.74)} & \cellcolor[HTML]{e9d0b2} AILS-NTUA (37.47) & \cellcolor[HTML]{e7cab2} ALPS-Lab (36.17) \\
Russian & Restaurant & \cellcolor[HTML]{dbedb2} PAI (55.99) & \cellcolor[HTML]{dfefb2} PALI (54.96) & \cellcolor[HTML]{f0f7b2} Takoyaki (51.30) & \cellcolor[HTML]{f2f8b2} \textbf{nchellwig (50.83)} & \cellcolor[HTML]{f3f9b2} ALPS-Lab (50.42) & \cellcolor[HTML]{f6fab2} TeamLasse (49.91) & \cellcolor[HTML]{f9f4b2} NLANGPROC (45.54) \\
Tatar & Restaurant & \cellcolor[HTML]{fdfcb2} Takoyaki (47.36) & \cellcolor[HTML]{faf4b2} \textbf{nchellwig (45.57)} & \cellcolor[HTML]{f9f2b2} PAI (45.23) & \cellcolor[HTML]{f7efb2} PALI (44.43) & \cellcolor[HTML]{f6edb2} ALPS-Lab (44.04) & \cellcolor[HTML]{f1e0b2} TeamLasse (41.13) & \cellcolor[HTML]{ead1b2} NLANGPROC (37.68) \\
Ukrainian & Restaurant & \cellcolor[HTML]{e2f0b2} PAI (54.37) & \cellcolor[HTML]{e7f3b2} PALI (53.07) & \cellcolor[HTML]{eef6b2} ALPS-Lab (51.63) & \cellcolor[HTML]{f4f9b2} Takoyaki (50.19) & \cellcolor[HTML]{fbfcb2} TeamLasse (48.79) & \cellcolor[HTML]{fdfcb2} \textbf{nchellwig (47.46)} & \cellcolor[HTML]{fbf7b2} NLANGPROC (46.31) \\
Chinese & Laptop & \cellcolor[HTML]{fdfeb2} NYCU Speech Lab (48.24) & \cellcolor[HTML]{f5e9b2} PALI (43.19) & \cellcolor[HTML]{f5e9b2} PAI (43.16) & \cellcolor[HTML]{efdcb2} \textbf{nchellwig (40.16)} & \cellcolor[HTML]{eedab2} ALPS-Lab (39.68) & \cellcolor[HTML]{ebd4b2} NLANGPROC (38.36) & \cellcolor[HTML]{e9d0b2} Takoyaki (37.45) \\
Chinese & Restaurant & \cellcolor[HTML]{deeeb2} NYCU Speech Lab (55.21) & \cellcolor[HTML]{e5f2b2} PAI (53.60) & \cellcolor[HTML]{e5f2b2} PALI (53.57) & \cellcolor[HTML]{f4f9b2} TeamLasse (50.26) & \cellcolor[HTML]{f7fbb2} \textbf{nchellwig (49.66)} & \cellcolor[HTML]{fcfdb2} ALPS-Lab (48.53) & \cellcolor[HTML]{fcf9b2} NLANGPROC (46.61) \\
\bottomrule
\end{tabular}
}
\caption{Subtask 3: \textsc{DimASQP}}
\label{tab:leaderboard-dimasqp}
\end{subtable}

\caption{Top-6 leaderboard rankings for the two subtasks across all language--domain pairs. Our submission is highlighted in \textbf{bold}.}
\label{tab:leaderboard_top5}
\end{table*}
%TODO Zeiten reporten

%% 1. Ergebnis Hyperparameter

Appendix~\ref{appendix:validation-performance} presents our validation results, showing that self-consistency prompting with $k=15$ generations yielded optimal average performance for DimASTE, whereas $k=10$ generations proved most effective for DimASQP. For the competition, we submitted the final test set predictions that can be observed when using these empirically determined values of $k$. 

\paragraph{Strong performance scores in the English language}
Table~\ref{tab:test_performance_combined} shows that English achieved the strongest overall performance across both subtasks, with the Restaurant domain outperforming Laptop for DimASTE and DimASQP. Across languages, we observe substantial performance differences: English consistently yields the highest cF1, while Tatar and Ukrainian are the most challenging languages.

\paragraph{Self-consistency improves test-set performance}
Across all language--domain pairs, SC improves average cF1 from 55.52 to 56.50 for DimASTE and from 46.10 to 47.37 for DimASQP. In terms of the underlying components, SC typically increases cPrec, while cRec is more stable and remains the limiting factor in several subsets (e.g., English--Laptop for both subtasks). Statistically significant improvements are marked in Table~\ref{tab:test_performance_combined} with symbols indicating the respective reference condition. Details on the significance testing procedure are provided in Appendix~\ref{appendix:significance}. Statistical analysis confirms that increasing the number of self-consistency views leads to widespread and frequently highly significant ($p < 0.001$) improvements in \textsc{cF1} across nearly all language--domain pairs. Higher view counts (5 or 10) also yield significant gains in some cases, especially for DimASQP.

%% Ergebnisse der Evaluation bei Leaderboard

\paragraph{Leaderboard}
Table~\ref{tab:leaderboard_top5} reports the official top-7 leaderboard rankings for each language--domain pair. For DimASTE, we ranked 2nd on English--Restaurant and 1st on Tatar--Restaurant, and stayed within the top six across all settings. For DimASQP, we ranked 2nd on both English subsets and remained competitive on the remaining languages. Our smallest gap to rank~1 was on English--Restaurant for DimASTE (69.85 vs.\ 70.21), while Chinese--Laptop showed the largest gap among the DimASTE subsets (40.16 vs.\ 48.24).

%% Trainingszeit reporten: Auf Tabelle im Appendix eingehen 2-3 Sätze.

\paragraph{PagedAttention yields substantial runtime gains}
Appendix~\ref{appendix:training_duration} quantifies the efficiency impact of vLLM's PagedAttention and batched inference. On average, evaluation without batching (w/o SC + w/o Batch.) is 26.6$\times$ slower than batched evaluation without SC for DimASTE (1.305 vs.\ 0.049\,s/ex.) and 28.2$\times$ slower for DimASQP (1.635 vs.\ 0.058\,s/ex.). This speed-up is key to making self-consistency feasible.

\paragraph{Ablation: LLM selection}\label{par:ablation-llms}
Appendix~\ref{appendix:llm-selection-performance} compares Gemma-3-27B, Mistral-Small-27B, and Qwen3-32B for different SC view counts. Overall, Mistral-Small achieves the best average performance for both subtasks (57.24 for DimASTE and 49.48 for DimASQP, both at 15 views). However, the best model varies by subset: for instance, Qwen3 is best on English--Laptop for DimASTE, while Gemma is best on Chinese--Restaurant for DimASTE. The results indicate that model choice interacts with language and domain, while SC consistently improves performance across LLMs.

% Weitere LLMs die auch auf mid range gpus laufen und wie gut diese waren berichten in 4-5 Sätzen

\section{Conclusion}

% Thema knapp zusammenfassen
% In Kontext verwandter Arbeiten setzen
% In Kontext leaderboard setzen
%% FW: LLM prompting
%% valence und arousal als phrase repräsentieren ? 
% Erwähnen, dass LLM abhängig
% FW: ein multilinguales Modell trainieren - mit mehr trainingsdaten ?
% FW: LLM Prompting evaluieren
% FW: Reasoning mit GRPO
% Limitation: viele leere predictions (?)
% Limitation: vor allem open source

We presented \textbf{\underline{S}elf-\underline{C}onsistent \underline{S}tructured \underline{G}eneration (SCSG)} for DimASTE and DimASQP. Our approach combines parameter-efficient fine-tuning with a self-consistency validation mechanism that aggregates multiple stochastic generations while averaging valence and arousal values for matching tuples. Across 6 languages and 8 language--domain combinations, self-consistency yields consistent gains on the official test sets, improving average cF1 from 55.52 to 56.50 for DimASTE and from 46.10 to 47.37 for DimASQP, with many improvements being statistically significant. Results vary strongly by language: English achieves the highest overall scores, whereas low-resource settings such as Tatar and Ukrainian remain substantially more challenging. On the SemEval leaderboard, SCSG achieved competitive rankings, including 2nd place on both English DimASQP subsets and 1st place on Tatar--Restaurant for DimASTE.

Future work could focus on several directions. First, one could systematically evaluate few-shot prompting variants to explore low-resource performance for DimABSA. Second, adapting multi-view prompting (MvP) as suggested by \citet{gou-etal-2023-mvp} for parameter-efficient fine-tuning of LLMs represents a promising avenue, which has thus far been applied exclusively to T5-based models. Notably, for MvP, training data scales by the number of possible sentiment element orderings (e.g., 24× for DimASQP), necessitating efficient strategies for managing computational overhead. Finally, one could investigate multilingual joint training by fine-tuning a single model on the combined data from all languages.

\newpage
\bibliography{custom}

\appendix
\onecolumn
\newpage

\section{Prompt}
\label{appendix:prompt}

\begin{figure*}[h]
    \centering
    \resizebox{1.0\columnwidth}{!}{%
    \includegraphics[page=1, width=\textwidth]{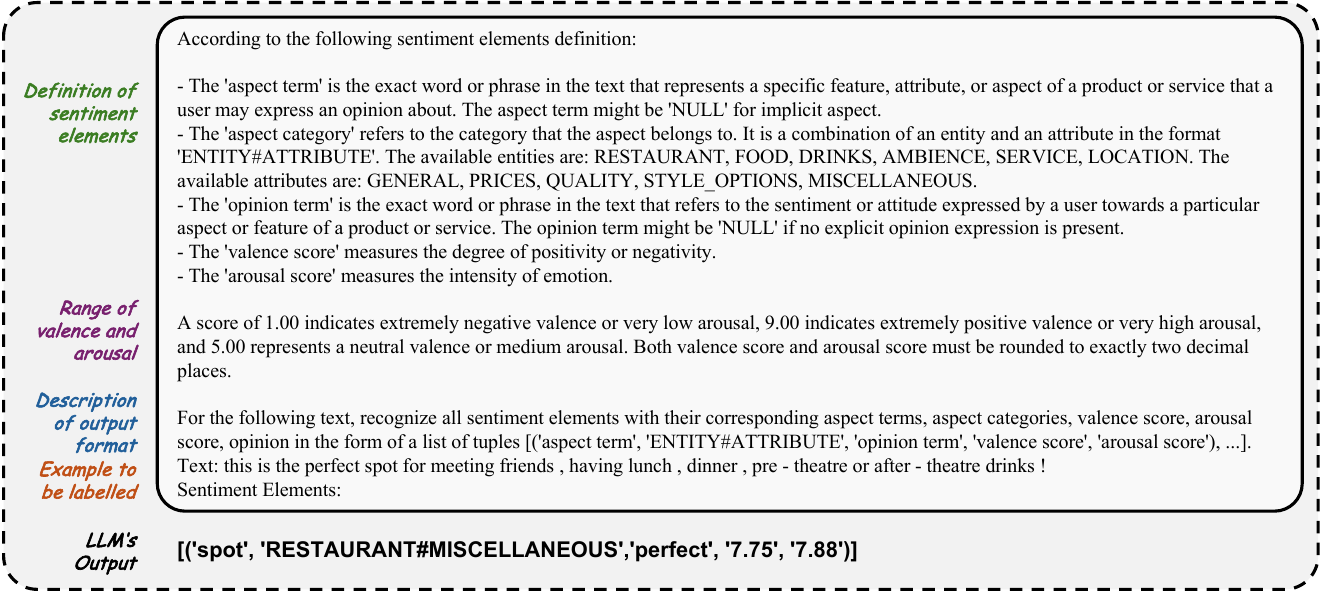}
    }
   \caption{Prompt used for SCSG. The prompt comprises descriptions of the considered sentiment elements (4 for DimASTE, 5 for DimASQP), explanations regarding the range of valence and arousal, the desired output format, and the example text for which ABSA is to be performed.}
    \label{fig:prompt-example}
\end{figure*}

\section{Datasets of SemEval 2026 Task 3: Subtask 2 \& 3 of Track A}
\label{appendix:dataset_stats}

\begin{table*}[h]
\centering
\small
\resizebox{1.0\columnwidth}{!}{%
\begin{tabular}{@{}llcccccc@{}}
\toprule
\textbf{Set} & \textbf{Domain} & \textbf{eng} & \textbf{jpn} & \textbf{rus} & \textbf{tat} & \textbf{ukr} & \textbf{zho} \\
\midrule
\multirow{3}{*}{\textbf{Train}} & Restaurant & 2{,}284 / 3{,}659 & --- & 1{,}240 / 2{,}487 & 1{,}240 / 2{,}487 & 1{,}240 / 2{,}487 & 6{,}050 / 8{,}523 \\
& Laptop     & 4{,}076 / 5{,}773 & --- & --- & --- & --- & 3{,}490 / 6{,}502 \\
& Hotel      & --- & 1{,}600 / 2{,}846 & --- & --- & --- & --- \\
\midrule
\multirow{3}{*}{\textbf{Dev}}   & Restaurant & 200 / 408 & --- & 48 / 102 & 48 / 102 & 48 / 102 & 300 / 761 \\
& Laptop     & 200 / 317 & --- & --- & --- & --- & 300 / 551 \\
& Hotel      & --- & 200 / 364 & --- & --- & --- & --- \\
\midrule
\multirow{3}{*}{\textbf{Test}}  & Restaurant & 1{,}000 / 2{,}129 & --- & 630 / 1{,}310 & 630 / 1{,}310 & 630 / 1{,}310 & 1{,}000 / 2{,}861 \\
& Laptop     & 1{,}000 / 1{,}975 & --- & --- & --- & --- & 1{,}000 / 2{,}798 \\
& Hotel      & --- & 800 / 1{,}443 & --- & --- & --- & --- \\
\bottomrule
\end{tabular}
}
\caption{Dataset statistics for DimASTE and DimASQP, reported as \emph{sentences / tuples} per split, language, and domain. A dash (---) indicates that no data is available for the respective language–domain combination.}
\label{tab:dataset_stats}
\end{table*}

\newpage

\section{Majority Voting Mechanism}\label{appendix:validation_mechanism}

\begin{figure*}[h]
\centering
\small
\definecolor{highlight1}{RGB}{230,242,255}
\definecolor{highlight2}{RGB}{235,248,235}
\definecolor{headerblue}{RGB}{240,248,255}
\setlength{\tabcolsep}{4pt}
\renewcommand{\arraystretch}{1.15}
\resizebox{\textwidth}{!}{%
\begin{tabular}{@{}clcclccc@{}}
\toprule
\multicolumn{8}{@{}l@{}}{\textit{Input:} ``Decor is nice though service can be spotty.''} \\
\midrule
\multicolumn{8}{@{}l@{}}{\textbf{Predictions}} \\
\cmidrule(r){1-1}\cmidrule(lr){2-4}\cmidrule(l){5-7}
Run & (Aspect Term, Sentiment Polarity) - Pair & Valence & Arousal & (Aspect Term, Sentiment Polarity) - Pair & Valence & Arousal & \\
\cmidrule(r){1-1}\cmidrule(lr){2-4}\cmidrule(l){5-7}
1 & \cellcolor{highlight1}(Decor, nice) & 6.92 & 7.13 & \cellcolor{highlight2}(service, spotty) & 5.53 & 6.03 & \\
2 & \cellcolor{highlight1}(Decor, nice) & 6.80 & 7.03 & (service, be spotty) & 5.60 & 6.10 & \\
3 & (Decor, is nice) & 6.67 & 6.90 & \cellcolor{highlight2}(service, spotty) & 5.40 & 5.90 & \\
4 & \cellcolor{highlight1}(Decor, nice) & 7.00 & 7.50 & \cellcolor{highlight2}(service, spotty) & 5.70 & 6.20 & \\
5 & (Decor, is nice) & 6.85 & 7.10 & \cellcolor{highlight2}(service, spotty) & 5.55 & 6.05 & \\
\midrule
\multicolumn{8}{@{}l@{}}{\textbf{Aggregation} (threshold $\tau = \lceil k/2 \rceil = 3$)} \\
\addlinespace[2pt]
\multicolumn{8}{@{}l@{}}{\textit{Tuple 1:} (Decor, nice) occurs in runs \{1, 2, 4\} $\rightarrow$ 3 occurrences $\geq \tau$} \\
\multicolumn{8}{@{}l@{}}{\quad\quad Valence: $\frac{6.92 + 6.80 + 7.00}{3} = 6.91$ \qquad Arousal: $\frac{7.13 + 7.03 + 7.50}{3} = 7.22$} \\
\addlinespace[3pt]
\multicolumn{8}{@{}l@{}}{\textit{Tuple 2:} (service, spotty) occurs in runs \{1, 3, 4, 5\} $\rightarrow$ 4 occurrences $\geq \tau$} \\
\multicolumn{8}{@{}l@{}}{\quad\quad Valence: $\frac{5.53 + 5.40 + 5.70 + 5.55}{4} = 5.54$ \qquad Arousal: $\frac{6.03 + 5.90 + 6.20 + 6.05}{4} = 6.04$} \\
\midrule
\multicolumn{8}{@{}l@{}}{\textbf{Final Output}} \\
\cmidrule(r){2-2}\cmidrule(lr){3-3}\cmidrule(l){4-4}
& Aspect-Sentiment Pair & Valence & Arousal & & & & \\
\cmidrule(r){2-2}\cmidrule(lr){3-3}\cmidrule(l){4-4}
& (Decor, nice) & 6.91 & 7.22 & & & & \\
& (service, spotty) & 5.54 & 6.04 & & & & \\
\bottomrule
\end{tabular}
}
\caption{Self-consistency majority voting for DimASTE over $k=5$ runs. Aspect-sentiment pairs (ignoring valence-arousal values) appearing in $\geq \tau = \lceil k/2 \rceil$ runs are aggregated by averaging their valence and arousal values. The aggregation section shows the explicit calculation. Light blue rows highlight matching (Decor, nice) variants; light green rows highlight matching (service, spotty) variants. For DimASQP, in addition to the aspect term and sentiment polarity, the aspect category is considered as well.}
\label{fig:majority_voting}
\end{figure*}

\section{Validation Performance}
\label{appendix:validation-performance}

\begin{table*}[h]
\centering
\label{tab:semeval_results}
\setlength{\tabcolsep}{10pt}
\resizebox{1.0\columnwidth}{!}{%
\begin{tabular}{@{}ll|cccc|cccc@{}}
\toprule
& & \multicolumn{4}{c|}{\textbf{\textsc{Subtask 2: DimASTE}}} & \multicolumn{4}{c}{\textsc{\textbf{Subtask 3: DimASQP}}} \\
\cmidrule(lr){3-6} \cmidrule(lr){7-10}
\textbf{Language} & \textbf{Domain} & BL & 5 Views & 10 Views & 15 Views & BL & 5 Views & 10 Views & 15 Views \\
\midrule
English & Restaurant & 77.93 & 78.35 & 78.15 & \textbf{78.45} & 75.17 & 75.30 & \textbf{75.39} & 75.26 \\
English & Laptop & 65.51 & \textbf{66.01} & 64.71 & 65.56 & 35.57 & 34.62 & \textbf{36.77} & 35.36 \\
Japanese & Hotel & 52.63 & \textbf{54.89} & 53.91 & 54.28 & 35.93 & \textbf{39.87} & 39.18 & 38.58 \\
Russian & Restaurant & 54.28 & 59.21 & 57.66 & \textbf{59.28} & 49.54 & \textbf{52.65} & 51.40 & 52.56 \\
Tatar & Restaurant & 52.72 & 52.83 & 52.99 & \textbf{53.54} & 38.65 & \textbf{44.65} & 43.95 & 43.56 \\
Ukrainian & Restaurant & 47.98 & 47.54 & 51.56 & \textbf{52.01} & 44.16 & 45.87 & \textbf{47.53} & 45.98 \\
Chinese & Restaurant & 65.12 & 65.05 & \textbf{65.73} & 65.47 & 58.77 & 60.90 & \textbf{60.95} & 60.78 \\
Chinese & Laptop & 45.15 & 45.30 & 45.40 & \textbf{45.62} & 36.73 & 37.50 & \textbf{38.33} & 38.15 \\
\midrule
\textbf{Average} &  & 57.67 & 58.65 & 58.76 & \textbf{59.28} & 46.82 & 48.92 & \textbf{49.19} & 48.78 \\
\bottomrule
\end{tabular}%
}
\caption{Evaluation on the development set for DimASTE and DimASQP: cF1 scores across languages and domains comparing vanilla prompting (BL) and self-consistency (SC) using either 5, 10 or 15 prompt executions, with majority voting applied. \textbf{Bold} values indicate the best performance for each language–domain pair.}
\end{table*}

\newpage

\section{Significance Testing}\label{appendix:significance}

For significance testing, we verified the normality of the $cF1$ scores across the five seeds using the Shapiro--Wilk test ($\alpha=0.05$) for each language--domain subset. Following the normality check, we first conducted an omnibus test (one-way ANOVA if normality held across all four groups, or Kruskal--Wallis test otherwise) as a gatekeeper, performing pairwise comparisons only if a globally significant difference ($p < 0.05$) was detected. Based on the result, we performed pairwise comparisons between all conditions using either two-sided independent $t$-tests (if normality holds) or two-sided Mann--Whitney $U$ tests (otherwise). To control the family-wise error rate, we corrected all $p$-values within each subtask using the Holm--Bonferroni correction ($\alpha=0.05$) \citep{holm1979simple}. 

\section{Training and Evaluation Duration}
\label{appendix:training_duration}

\begin{table*}[h]
\centering
\small

\resizebox{1.0\columnwidth}{!}{%
\begin{tabular}{llrrrrrrrrrr}
\toprule
 & & \multicolumn{3}{c}{\textbf{Training}} & \multicolumn{7}{c}{\textbf{Evaluation}} \\
\cmidrule(lr){3-5} \cmidrule(lr){6-12}
\textbf{Language} & \textbf{Domain} & \textbf{\#} & \textbf{Time (s)} & \textbf{Time/1K (s)} & \textbf{\#} & \multicolumn{2}{c}{\textbf{w/ SC}} & \multicolumn{2}{c}{\textbf{w/o SC}} & \multicolumn{2}{c}{\textbf{w/o SC + w/o Batch.}} \\
\cmidrule(lr){7-8} \cmidrule(lr){9-10} \cmidrule(lr){11-12}
 & & & & & & \textbf{Time (s)} & \textbf{Time/ex. (s)} & \textbf{Time (s)} & \textbf{Time/ex. (s)} & \textbf{Time (s)} & \textbf{Time/ex. (s)} \\
\midrule
English & Restaurant & 2,284 & 3,860 & 1,690 & 1,000 & 320 & 0.321 & 37 & 0.037  & 1,151 & 1.151  \\
English & Laptop & 4,076 & 6,515 & 1,598 & 1,000 & 289 & 0.289 & 32 & 0.032  & 1,017 & 1.018  \\
Japanese & Hotel & 1,600 & 3,415 & 2,134 & 800 & 234 & 0.293 & 29 & 0.037  & 822 & 1.028  \\
Russian & Restaurant & 1,240 & 2,638 & 2,127 & 630 & 277 & 0.440 & 38 & 0.060  & 884 & 1.405  \\
Tatar & Restaurant & 1,240 & 3,686 & 2,972 & 630 & 303 & 0.481 & 40 & 0.065  & 924 & 1.467  \\
Ukrainian & Restaurant & 1,240 & 2,935 & 2,367 & 630 & 287 & 0.456 & 40 & 0.065  & 908 & 1.442  \\
Chinese & Restaurant & 6,050 & 10,281 & 1,699 & 1,000 & 493 & 0.493 & 52 & 0.053  & 1,687 & 1.688  \\
Chinese & Laptop & 3,490 & 7,143 & 2,046 & 1,000 & 361 & 0.362 & 42 & 0.042  & 1,240 & 1.240  \\
\midrule
\textbf{Average} &   & 2,652 & 5,059 & 2,079 & 836 & 320 & 0.392 & 39 & 0.049 & 1,079 & 1.305  \\
\bottomrule
\end{tabular}
}

\vspace{0.2em}
(a) Subtask 2: \textsc{DimASTE}

\vspace{1.5em}

\resizebox{1.0\columnwidth}{!}{%
\begin{tabular}{llrrrrrrrrrr}
\toprule
 & & \multicolumn{3}{c}{\textbf{Training}} & \multicolumn{7}{c}{\textbf{Evaluation}} \\
\cmidrule(lr){3-5} \cmidrule(lr){6-12}
\textbf{Language} & \textbf{Domain} & \textbf{\#} & \textbf{Time (s)} & \textbf{Time/1K (s)} & \textbf{\#} & \multicolumn{2}{c}{\textbf{w/ SC}} & \multicolumn{2}{c}{\textbf{w/o SC}} & \multicolumn{2}{c}{\textbf{w/o SC + w/o Batch.}} \\
\cmidrule(lr){7-8} \cmidrule(lr){9-10} \cmidrule(lr){11-12}
 & & & & & & \textbf{Time (s)} & \textbf{Time/ex. (s)} & \textbf{Time (s)} & \textbf{Time/ex. (s)} & \textbf{Time (s)} & \textbf{Time/ex. (s)} \\
\midrule
English & Restaurant & 2,284 & 4,957 & 2,170 & 1,000 & 403 & 0.403 & 41 & 0.042  & 1,417 & 1.417  \\
English & Laptop & 4,076 & 9,678 & 2,374 & 1,000 & 397 & 0.398 & 42 & 0.042  & 1,370 & 1.371  \\
Japanese & Hotel & 1,600 & 4,496 & 2,810 & 800 & 316 & 0.395 & 36 & 0.046  & 1,090 & 1.364  \\
Russian & Restaurant & 1,240 & 3,266 & 2,633 & 630 & 312 & 0.497 & 45 & 0.072  & 1,046 & 1.662  \\
Tatar & Restaurant & 1,240 & 4,489 & 3,620 & 630 & 343 & 0.545 & 46 & 0.074  & 1,120 & 1.778  \\
Ukrainian & Restaurant & 1,240 & 3,636 & 2,932 & 630 & 354 & 0.562 & 45 & 0.073  & 1,107 & 1.758  \\
Chinese & Restaurant & 6,050 & 12,829 & 2,120 & 1,000 & 625 & 0.626 & 62 & 0.062  & 2,121 & 2.121  \\
Chinese & Laptop & 3,490 & 10,066 & 2,884 & 1,000 & 473 & 0.473 & 52 & 0.052  & 1,610 & 1.611  \\
\midrule
\textbf{Average} &   & 2,652 & 6,677 & 2,693 & 836 & 403 & 0.487 & 46 & 0.058 & 1,360 & 1.635  \\
\bottomrule
\end{tabular}
}

\vspace{0.2em}
(b) Subtask 3: \textsc{DimASQP}

\caption{Training and evaluation duration across languages and domains for both DimASTE and DimASQP. We report duration with Self-Consistency (w/ SC, 15 views), without (w/o SC), and without vLLM's batch processing (w/o Batch.).}
\label{tab:dataset_duration}
\end{table*}

\newpage
\section{Performance Comparison Across LLMs}
\label{appendix:llm-selection-performance}

\begin{table*}[h]
    \centering
    \small
    \begin{subtable}{\textwidth}
        \centering
        \resizebox{\textwidth}{!}{
        \begin{tabular}{ll ccc ccc ccc ccc}
            \toprule
            \multirow{2}{*}{\textbf{Language}} & \multirow{2}{*}{\textbf{Domain}} & \multicolumn{3}{c}{\textbf{Baseline}} & \multicolumn{3}{c}{\textbf{5 Views}} & \multicolumn{3}{c}{\textbf{10 Views}} & \multicolumn{3}{c}{\textbf{15 Views}} \\
            \cmidrule(lr){3-5} \cmidrule(lr){6-8} \cmidrule(lr){9-11} \cmidrule(lr){12-14}
            & & G & M & Q & G & M & Q & G & M & Q & G & M & Q \\
            \midrule
English & Laptop & 60.48 & 61.44 & 63.05 & 60.64 & 61.07 & 63.61 & 60.99 & 61.04 & 63.56 & 60.87 & 61.18 & \textbf{63.66} \\
English & Restaurant & 69.70 & 69.09 & 67.40 & 69.85 & 69.60 & 67.82 & \textbf{70.13} & 69.66 & 67.89 & 69.92 & 70.01 & 68.23 \\
Japanese & Hotel & 53.96 & 56.64 & 49.33 & 54.46 & 56.83 & 49.97 & 55.16 & 56.73 & 49.81 & 55.35 & \textbf{57.91} & 50.38 \\
Russian & Restaurant & 55.24 & 55.78 & 52.63 & 56.13 & 57.36 & 53.45 & 55.88 & 57.39 & 53.69 & 56.40 & \textbf{57.45} & 54.74 \\
Tatar & Restaurant & 48.94 & 49.63 & 43.87 & 49.96 & \cellcolor[HTML]{C8E6C9} 51.26 & 45.15 & 50.42 & \cellcolor[HTML]{C8E6C9} 51.37 & 45.92 & 51.00 & \cellcolor[HTML]{C8E6C9} \textbf{51.65} & 46.21 \\
Ukrainian & Restaurant & 51.82 & 52.85 & 49.77 & 52.89 & 54.43 & 50.74 & 52.77 & 54.55 & 50.42 & 52.69 & \textbf{54.87} & 51.17 \\
Chinese & Laptop & 49.13 & 50.33 & 47.48 & 50.20 & 51.25 & 47.83 & 50.61 & 51.31 & 48.50 & 50.86 & \textbf{51.52} & 48.75 \\
Chinese & Restaurant & 54.87 & 52.80 & 52.07 & 54.51 & 53.28 & 53.21 & 54.86 & 52.92 & 53.49 & \textbf{54.90} & 53.35 & 53.30 \\
\midrule
\multicolumn{2}{l}{\textbf{\textbf{Average}}} & 55.52 & 56.07 & 53.20 & 56.08 & 56.88 & 53.97 & 56.35 & 56.87 & 54.16 & 56.50 & \textbf{57.24} & 54.55 \\
\bottomrule
        \end{tabular}
        }
        \subcaption{Subtask 2: \textsc{DimASTE}}
    \end{subtable}

    \vspace{1.5em}

    \begin{subtable}{\textwidth}
        \centering
        \resizebox{\textwidth}{!}{
        \begin{tabular}{ll ccc ccc ccc ccc}
            \toprule
            \multirow{2}{*}{\textbf{Language}} & \multirow{2}{*}{\textbf{Domain}} & \multicolumn{3}{c}{\textbf{Baseline}} & \multicolumn{3}{c}{\textbf{5 Views}} & \multicolumn{3}{c}{\textbf{10 Views}} & \multicolumn{3}{c}{\textbf{15 Views}} \\
            \cmidrule(lr){3-5} \cmidrule(lr){6-8} \cmidrule(lr){9-11} \cmidrule(lr){12-14}
            & & G & M & Q & G & M & Q & G & M & Q & G & M & Q \\
            \midrule
English & Laptop & 39.43 & 41.39 & 38.96 & 39.39 & \textbf{42.25} & 39.05 & 40.11 & 41.80 & 38.56 & 40.27 & 42.12 & 39.56 \\
English & Restaurant & 63.27 & 64.90 & 60.62 & 63.59 & \cellcolor[HTML]{C8E6C9} 65.60 & 61.83 & 63.88 & \cellcolor[HTML]{C8E6C9} 65.39 & 61.69 & 63.97 & \cellcolor[HTML]{C8E6C9} \textbf{65.93} & 61.81 \\
Japanese & Hotel & 38.23 & 42.16 & 34.24 & 39.20 & \cellcolor[HTML]{C8E6C9} 43.74 & 33.54 & 40.06 & \cellcolor[HTML]{C8E6C9} 44.41 & 34.28 & 40.37 & \cellcolor[HTML]{C8E6C9} \textbf{44.78} & 34.46 \\
Russian & Restaurant & 49.69 & 50.99 & 46.64 & 50.64 & 52.78 & 48.91 & 50.83 & 52.50 & 49.06 & 50.96 & \textbf{52.87} & 49.12 \\
Tatar & Restaurant & 44.49 & 43.24 & 37.51 & 46.12 & \textbf{46.51} & 40.30 & 45.85 & 45.93 & 40.14 & 45.90 & 46.17 & 40.94 \\
Ukrainian & Restaurant & 45.90 & 48.35 & 42.83 & 46.84 & 49.17 & 44.16 & 47.29 & 50.08 & 44.61 & 47.59 & \textbf{50.50} & 45.05 \\
Chinese & Laptop & 39.10 & 41.44 & 36.54 & 39.71 & 42.96 & 38.58 & 40.07 & 43.53 & 39.11 & 40.24 & \textbf{43.60} & 39.41 \\
Chinese & Restaurant & 48.66 & 47.82 & 44.68 & 49.52 & 48.83 & 45.33 & 49.75 & 49.21 & 45.99 & 49.68 & \textbf{49.87} & 46.15 \\
\midrule
\multicolumn{2}{l}{\textbf{\textbf{Average}}} & 46.10 & 47.54 & 42.75 & 46.88 & 48.98 & 43.96 & 47.23 & 49.11 & 44.18 & 47.37 & \textbf{49.48} & 44.56 \\
\bottomrule
        \end{tabular}
        }
        \subcaption{Subtask 3: \textsc{DimASQP}}
    \end{subtable}
   \caption{Evaluation on different LLMs (cF1). We compared \textbf{G} (Gemma-3-27B), \textbf{M} (Mistral-Small-24B), and \textbf{Q} (Qwen3-32B). Green cells (\colorbox[HTML]{C8E6C9}{\phantom{x}}) indicate performance surpassing the Rank 1 team on the official competition leaderboard for the respective language and domain combination. Overall, the results demonstrate that increasing the number of self-consistency views leads to consistent performance gains, with several configurations significantly outperforming the competition's top-ranked entries.}
   \label{tab:final_results}
\end{table*}

\end{document}